\documentclass{article}
\usepackage{spconf,amsmath,graphicx}
\usepackage{graphicx}
\usepackage{times}
\usepackage{helvet}
\usepackage{courier}
\usepackage{amsmath}
\usepackage{csquotes}
\usepackage{color}
\usepackage{paralist}
\usepackage{amssymb}
\usepackage{amsthm}
\usepackage{indentfirst}
\usepackage{subfigure}
\usepackage{float}
\usepackage{multirow}
\usepackage{cite}
\usepackage{hyperref}
\usepackage{mathrsfs}
\title{Learning Target-oriented Dual Attention for Robust RGB-T Tracking}
\name{Rui Yang, Yabin Zhu, Xiao Wang, Chenglong Li, Jin Tang}
\address{Anhui University \\
	School of Computer Science and Technology \\
	Jiulong Road No. 111, Hefei, Anhui Province, China}

\begin{document}
%\ninept
%
\maketitle

\begin{abstract}
RGB-Thermal object tracking attempt to locate target object using complementary visual and thermal infrared data. Existing RGB-T trackers fuse different modalities by robust feature representation learning or adaptive modal weighting. However, how to integrate dual attention mechanism for visual tracking is still a subject that has not been studied yet. In this paper, we propose two visual attention mechanisms for robust RGB-T object tracking. Specifically, the local attention is implemented by exploiting the common visual attention of RGB and thermal data to train deep classifiers. We also introduce the global attention, which is a multi-modal target-driven attention estimation network. It can provide global proposals for the classifier together with local proposals extracted from previous tracking result. Extensive experiments on two RGB-T benchmark datasets validated the effectiveness of our proposed algorithm.
\end{abstract}


\section{Introduction}
Visual tracking has achieved great success in recent years due to the wide application of neural networks \cite{wang2018sint++, xiong2018learning}. However, tracking performance in the wild scenario is still unsatisfactory, due to the influence of illumination, clutter background, heavy occlusion and out-of-view, \emph{etc.}

Recently, some researchers resort to other data to help improve the robustness of visual trackers, such as thermal images \cite{li2018rgb}, natural language description \cite{wang2018describe} and depth images \cite{song2013tracking}. Compared to text and depth images, thermal sensor is not sensitive to lighting condition and can capture the target object at far distance, and it still works well at night while RGB, depth or text may failed. What's more, it has a strong ability to penetrate haze as well as smog. Therefore, RGB-T tracking receives much attention in recent years. However, the tracking performance of existing RGB-T algorithms is also unsatisfactory to some extent due to aforementioned challenging factors. 
	
Visual attention has a great potential of facilitating learning discriminative classifiers for RGB-T tracking. Existing deep attentive trackers \cite{zhu2018fanet, chu2017online} mainly adopt additional attention modules for estimating feature weights to strengthen the discriminative power of features and improve the tracking accuracy. However, the feature weights learned in single frame may not enable classifier to concentrate on robust features over a long temporal span. Moreover, slight inaccuracy of feature weights will exacerbate the misclassification problem. Similar views can also be found in DAT tracker \cite{pu2018deep}. This inspired us to think how the visual attention can help the tracker attend to target objects over time? 

In this paper, we propose a novel dual visual attention guided deep RGB-T tracking algorithm, \emph{i.e.} the local attention and global attention. The training process consists of both a forward and a backward step. In the forward step, we feed the paired RGB and thermal samples into a deep tracking-by-detection network and estimate the corresponding classification score. In the backward step, we take the partial derivative of this classification score with respect to the input paired RGB-T samples along the direction from the last fully connected layer towards the first convolutional layer. We take the partial derivative output of the first layer as the common attention map of RGB and thermal inputs. Each pixel value on this attention map indicates the importance of the corresponding pixels of the input RGB-T samples to affect the classification accuracy. We add this attention map in the loss function as a regularization term during training to make the classifier pay more attention to target regions.

Although aforementioned RGB-T tracker can already achieve good performance, however, it still follows the local search strategy under tracking-by-detection framework. This will make the tracker sensitive to challenging factors, due to previous tracking result may already invalid for the ensure of candidate search window. Similar views can also be found in \cite{wang2018describe, zhu2016beyond}. Therefore, we extend the target-driven attention estimation network which is first proposed in \cite{wang2018describe} into the RGB-T global attention version to handle the issues caused by local search strategy. Specifically, we take the RGB, thermal images and initial target objects as inputs, and concatenate the feature maps extracted from fully convolutional network. These features are fed into an up-sample network to generate corresponding attention map. High quality global proposals can be extracted from the attention regions and fed into the classifier together with local proposals. Hence, the complementary of local and global attention maps will further improve the robustness and accuracy of RGB-T object tracker.

The contributions of this paper can be summarized as the following three aspects:
1) We propose a local attention mechanism to exploit visual attention for RGB-T object tracking.
2) We extend the target-driven global attention mechanism into multi-modal version to further improve the robustness of RGB-T tracker.
3) We conduct extensive experiments on two RGB-T benchmark datasets to validate the effectiveness of our tracker and also the attention modules.

\section{Related Works}

\textbf{RGB-T Visual Tracking.}
RGB-T tracking receives more and more attention in computer vision community with the popularity of thermal infrared sensors. Wu \emph{et al.} ~\cite{Wu2011Multiple} concatenate the image patches from RGB and thermal sources, and then sparsely represent each sample in the target template space for tracking. Modal weights are introduced for each source to represent the image quality, and combine with  the sparse representation in Bayesian filtering framework to perform object tracking~\cite{li2017weighted}. Zhu \emph{et al.} ~\cite{zhu2018fanet} propose a quality-aware feature aggregation network to fuse multi-layer deep feature and multimodal information adaptively for RGB-T tracking. Although these works also achieve good performance on RGB-T benchmarks, however, they still adopt local search strategy for target localization and seldom of them consider explore long-term visual attentions for their tracker. 

\textbf{Attention for Visual Tracking.}
Attention mechanism originates from the study of human congnitive neuroscience ~\cite{Olshausen1993A}. In visual tracking, the cosine window map ~\cite{bolme2010visual} and Gaussian window map ~\cite{danelljan2015learning} are widely used in DCF tracker to suppress the boundary effect, which can interpreted as one type of visual spatial attention.  For short-time tracking, DAVT ~\cite{fan2010discriminative} used a discriminative spatial attention and  ACFN ~\cite{choi2017attentional} developed an attentional mechanism that chose a subset of the associated correlation filters for visual tracking. Recently, RASNet ~\cite{wang2018learning} integrates the spatial attention, channel attention and residual attention to achieve the state-of-the-art tracking accuracy. Different from these works, we propose a novel local and global attention for robust RGB-T tracking.

\section{Our Tracker}

\subsection{Network Architecture}
\begin{figure}[t]
\center
\includegraphics[width=3.3in]{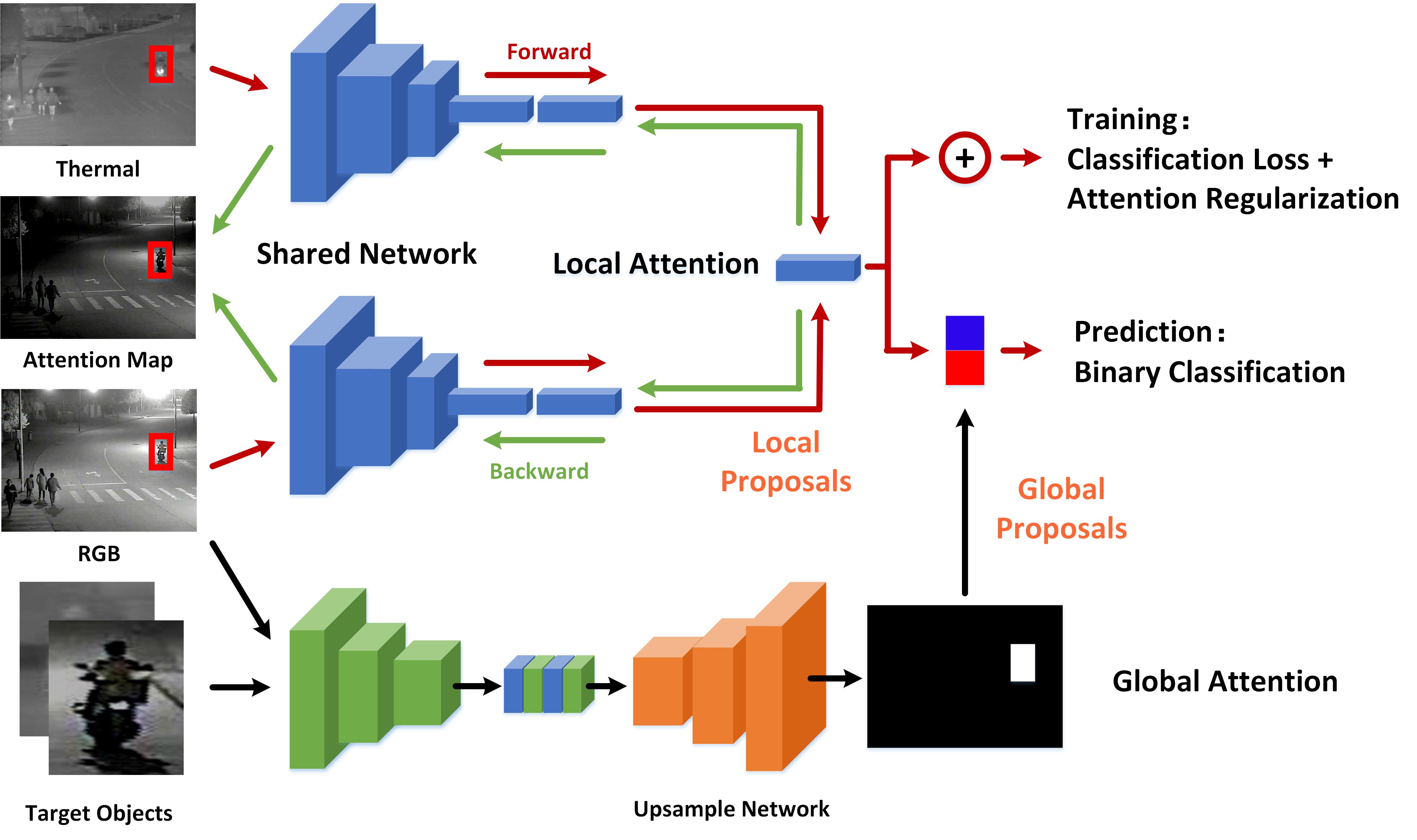}
\caption{The pipeline of our proposed tracking algorithm. }
\label{pipeline}
\end{figure}

The overall pipeline of our RGB-T tracker can be found in Fig. \ref{pipeline}. Our tracker contains two main modules, \emph{i.e.}, local attention based RGB-T tracker and multi-modal target-driven global attention estimation module. 
	
\textbf{Local Attention Network. }	The regular tracking-by-detection framework usually defines the target object as the positive class and the background as the negative class to train a binary classifier, such as MDNet \cite{nam2016learning}. We adopt MDNet as the core of our RGB-T baseline tracker due to its powerful feature representation. Specifically, for the input RGB and thermal samples pairs, we use three convolutional layers and two fully connected layers to extract their features. To reduce the computational burden, the features of different models are concatenated and fed into the domain-specific layers to get the score map.
The cross entropy loss is used for the optimization, which can be written as follows:
\begin{equation}
\label{entropyLoss}
L_c = -\sum_{i=1}^{N}y_i log P_i + (1- y_i) log(1- P_i)
\end{equation}
where $N$ is the mini-batch size, $y_i$ the ground truth label of the $i$-th RGB-T sample pair, $P_i$ is the prediction of corresponding RGB-T sample pair.

To make the classifier focuses more attention on the target object in the tracking procedure, we add a regularization term based on the cross-entropy function used in MDNet. The motivation of this term is: we can obtain two attention maps for the input pairs, \emph{i.e.}, the positive attention map $A_p$ and the negative attention map $A_n$. For each positive sample, we expect the pixel values of $A_p$ related to target objects to be large and the pixel values $A_n$ should be small. Therefore, the attention regularization term for each positive sample can be written as:
\begin{equation}
\label{posAttentionRegularization}
R_{(y=1)} = \frac{\sigma_{A_p}}{\mu_{A_p}} + \frac{\mu_{A_n}}{\sigma_{A_n}},
\end{equation}
where $\mu$ and $\sigma$ are the mean and standard deviation operators for the attention maps. We can also obtain corresponding regularization term for the negative training samples as:
\begin{equation}
\label{negAttentionRegularization}
R_{(y=0)} = \frac{\mu_{A_p}}{\sigma_{A_p}} + \frac{\sigma_{A_n}}{\mu_{A_n}},
\end{equation}
The final formulation of our classification and regularization term can be written as:
\begin{equation}
\label{finalformulation}
L = L_c + \lambda * [y * R_{y=1} + (1-y) * R_{y=0}],
\end{equation}
where $R_{y=1}$ and $R_{y=0}$ denote the regularization terms of the positive and negative training examples, respectively. $\lambda$ is a scalar parameter used to balance the two terms. In our experiments, we also check the influence of this parameter.  

Based on Eq. \ref{finalformulation}, we can conduct the reciprocative learning \cite{pu2018deep} via standard backward propagation and chain rule. The attention maps of each input training data can be obtained in each iteration of the classifier training. The classifier will pay more attention to target objects than the background. In the tracking stage, the attention terms will not be used and the classifier will learn to attend the RGB and thermal image pairs. 

Although we can already obtain good performance with the local attention mechanism, however, this improved tracking-by-detection framework still adopt the local search strategy which may lead to sensitive to heavy occlusion, out-of-view and abrupt motion, \emph{etc}. Similar views can also be found in \cite{wang2018describe, zhu2016beyond}. Hence, we introduce the RGB-T target-driven global attention network to handle this issue. The detailed introduction can be found in the following subsection.

\textbf{Global Attention Network.} In this subsection, we propose the RGB-T target-driven global attention estimation network to complement with local proposals for robust visual tracking. As shown in Fig. \ref{pipeline}, the input of this module are RGB, thermal video frame and corresponding target objects. We adopt the truncated VGG network to extract the feature representation of these inputs and concatenate into one feature maps. Specifically, we first resize all the input image and patches into $192 \times 256 \times 3$ and the corresponding feature map is $12 \times 16 \times 512$. Therefore, the concatenated feature map is $12 \times 16 \times 2048$ and then fed into the upsample network. The upsample network is a reversed VGG network and its output has the same resolution with the input image.

\subsection{Training}
Different from regular deep trackers which may need to pre-training on large scale classification dataset, our proposed local attention based RGB-T tracker is not need such pre-training. Following \cite{pu2018deep}, we only initialize the parameters of our network using VGG network which only pre-trained on ImageNet dataset for the classification task. Our experiments validate that our tracker can achieve comparable or even better tracking performance on RGB-T dataset with the help of local attention mechanism.

For the RGB-T target-driven global attention network, we use GTOT-50 \cite{li2017grayscale} as training dataset and tracking on RGBT-234 dataset \cite{li2018rgb}, and also use RGBT-234 dataset for training and test on GTOT-50 dataset. Following \cite{wang2018describe}, the ground truth binary mask used for training is generated from corresponding training dataset. Specifically, we first generate a black mask which has the same resolution as the video frame, then, we bleach the target object regions according to annotated BBox in the training dataset.

\subsection{Tracking}
In the first frame, positive and negative samples are extracted around the initial target location according to their intersection  over union (IoU) with ground truth BBox. The samples will be regarded as positive samples if their IoU in range of 0.7 to 1 and negative samples if the overlap is less than 0.5.

For each video frame, we first draw $N$ proposals around previous tracking result and global attention regions. Then these samples will be fed into our RGB-T tracking network. The proposal with maximum classification score will be chosen as the candidate result of current frame. After that, we adopt bounding box regression to refine the target location.

\section{Experiments}

\subsection{Datasets and Evaluation Criteria}
The GTOT-50 and RGBT-234 datasets are used in our experiment to validate the effectiveness of our tracker. GTOT-50 contains 50 RGB-T video sequences including more than six hundred RGB-T matching image pairs. RGBT-234 contains 234 RGB-T video sequences including more than 11000 RGB-T matching image pairs totally. 
In our experiments, we use precision rate (PR) and success rate (SR) as the evaluation criteria of algorithm performance. The threshold of distance are set to 5 pixels for GTOT-50 and RGBT-234 equally, and the threshold for overlap is set to 0.6 for both datasets. 

\begin{figure}[t]
\centering
\includegraphics[width=3.3in]{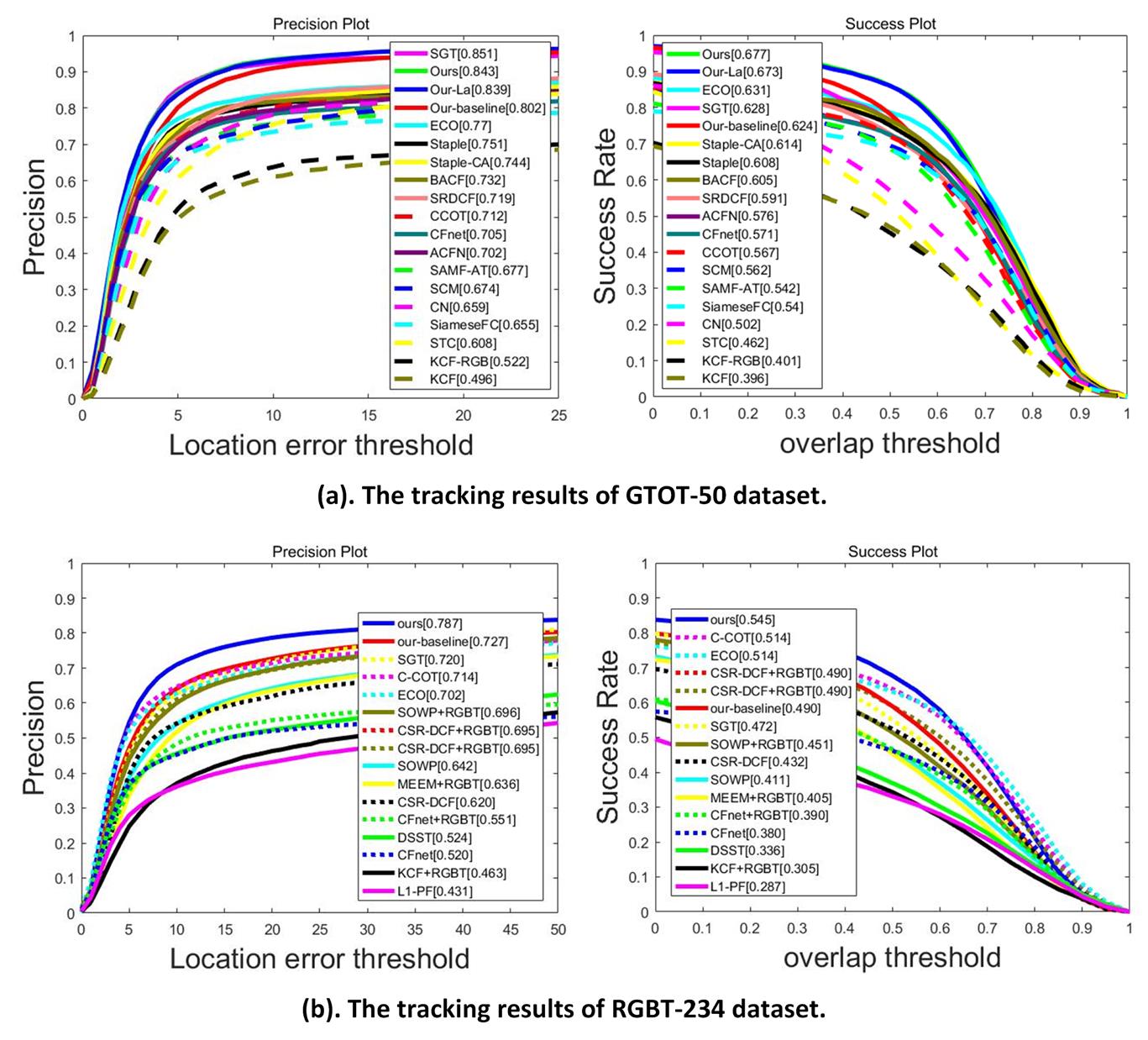}
\caption{ Tracking results on GTOT-50 and RGBT-234 dataset. }
\label{trackingResults}
\end{figure}

\begin{figure}[t]
\centering
\includegraphics[width=3.3in]{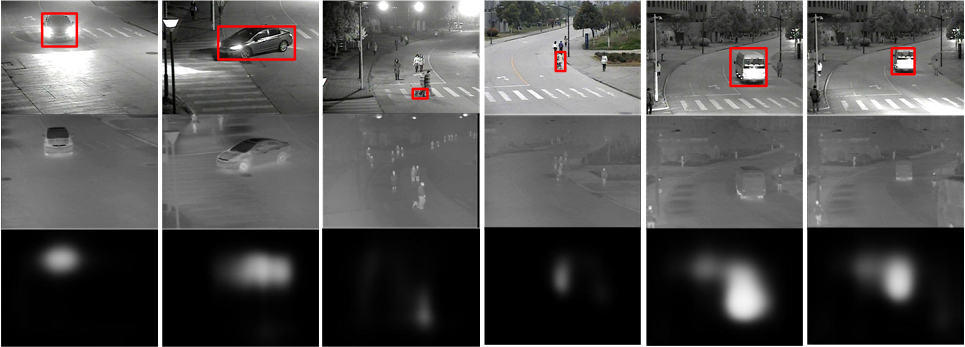}
\caption{ Attention maps generated by our attention network. }
\label{globalAttentionResults}
\end{figure}

\subsection{Implementation Details}
Due to the insufficient images in GTOT-50 dataset, it is hard to train a robust global attention generator. Therefore, we train the attention network based on both GTOT-50 and LaSOT dataset \cite{Fan2018LaSOT} (we replace the thermal data with RGB image for the LaSOT dataset). We set the target object pixels as zero and background pixels as 255 to obtain binary mask for each video frame. Those masks are used as ground truth attention maps to optimize the global attention network. It is also worthy to note that, only select 44660 images are selected from the LaSOT dataset for the training to quickly validate our proposed method (original dataset contains 3.52 million frames). To avoid the influence of inaccurate global proposals, we propose some constraints to filter out them: If the distance between the center of global proposals and the center of the result in the previous frame exceeds a pre-setting scalar or the overlap between them is less than the threshold, the global proposals will be discarded. We experimentally set the distance and overlap threshold as 25 and 0.3, respectively.

The initial learning rate is 5e-5, batchsize is 20, Adagrad is used for the optimization. The network is trained for 50 epoches. All the experiments are implemented based on PyTorch on a desktop computer with Ubuntu 16.04, I7-6700K, NVIDIA GTX-1080TI with 11G VRAM and 32G RAM.

\subsection{Comparison with other trackers}
\begin{figure*}[t]
\center
\includegraphics[width=7in]{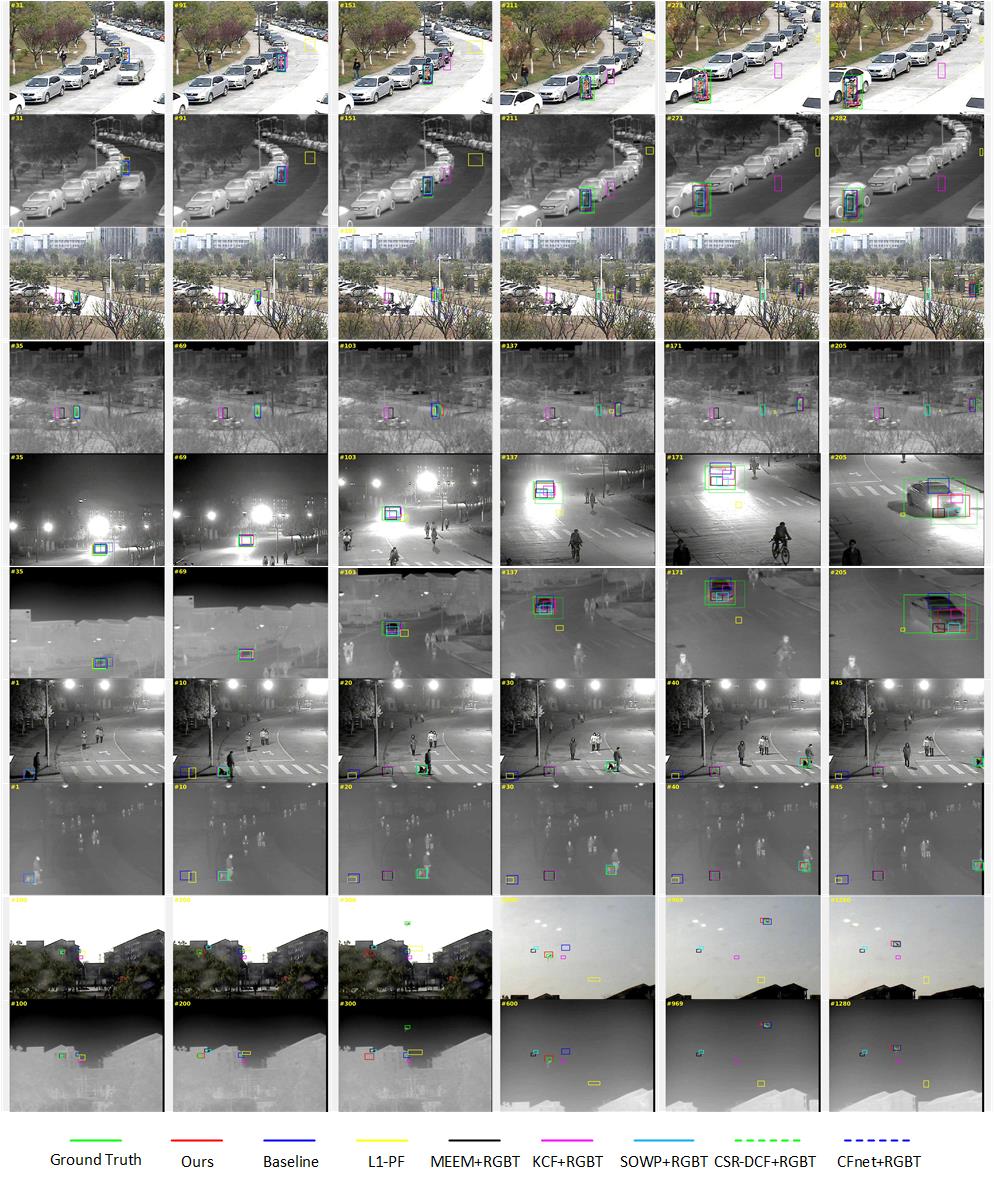}
\caption{Results of our tracker and other tracking algorithm.}
\label{visualizationResults}
\end{figure*}

For the RGBT-234 dataset, the overall tracking results can be found in Fig. \ref{trackingResults}. Our tracker obtains 0.545 and 0.787 on SR and PR criterion which are all significantly better than other trackers. This fully demonstrate the effectiveness of our proposed RGB-T tracker. Moreover, we visualize some attention maps and  tracking results in Fig. \ref{globalAttentionResults} and Fig. \ref{visualizationResults}, respectively.

As shown in Fig. \ref{trackingResults}, our tracker achieves the state-of-the-art performance on SR and also second top result on PR criterion on GTOT-50 dataset when compared with other tracking algorithms. Specifically, our tracker achieves 0.677 on SR which is significantly better than second top algorithm ECO (SR: 0.631). This also fully demonstrate the effectiveness of our proposed RGB-T tracker. However, our tracker is not as good as SGT on this dataset on PR criterion, we think this maybe caused by the utilization of inter-modal weighting and patches weighting mechanism of SGT and the bias caused by the smaller GTOT-50 dataset. In our future works, we will consider to introduce co-attention and multi-scale mechanism to achieve better tracking results.

\subsection{Ablation Study}
To demonstrate the significance of our proposed method, we conduct experiments of component analysis on GTOT-50 dataset.
The following three models are designed to check the effectiveness of each component:
1) Our-baseline, we remove both the reciprocative learning algorithm and target-driven attention mechanism.
2) Our-La, only the reciprocative learning algorithm is used.
3) Ours, both reciprocative learning and target-driven attention are all used.

The experimental results are shown in Fig. \ref{trackingResults}. According to the results, the PR and SR of Our-La is 3.7 and 4.9 points higher than the baseline respectively. This fully validate the effectiveness of our multi-modal  reciprocative learning algorithm. In addition, we can find that the target-driven global proposal mechanism improves 0.4 points on both PR and SR criteria. This also proved that the global attention could also help improve the results of visual trackers.

To validate the influence of different values of $\lambda$ in Eq. \ref{finalformulation}, we conduct following experiment to check the final tracking results. As shown in Table \ref{result-nmeta}, we set the $\lambda$ from 1 to 9, and the tracking results is nearly the same. Therefore, we can find that our RGB-T tracker is not sensitive to this parameter.

\begin{table}[htp!]
\center
\scriptsize
\caption{The tracking results with different values of $\lambda$. }\label{result-nmeta}
\begin{tabular}{l|ccccccccr}
\hline
\hline
$\lambda$   	   &1   &2 &3 &4 &5 &6 &7 &8  &9 \\
\hline															
PR 			            &84.3 	&84.9  &84.3   &84.8  &83.5  &84.1  &83.8	&84.6 &	84.7			\\
\hline	
SR 			            &67.4 	&67.4  &67.7&67.4  &66.9  &67.5  &67.1 &66.9 &	67.7		\\
\hline	
\end{tabular}
\end{table}

In addition, we also report the running time of our tracker and baseline tracker: our tracker achieves 0.349 FPS while the baseline tracker achieves 0.633 FPS and DAT tracker (RGBT version) is 0.465 FPS. Although our tracker is a little slower than the baseline tracker, however, we can obtains better tracking results and more robust to challenging factors. Besides, the proposed dual-attention mechanism is generic and can also be integrated with real-time tracker, such as RT-MDNet \cite{Jung_RTMDNET}. We leave this as our future works.

\begin{figure*}[t]
\center
\includegraphics[width=7in]{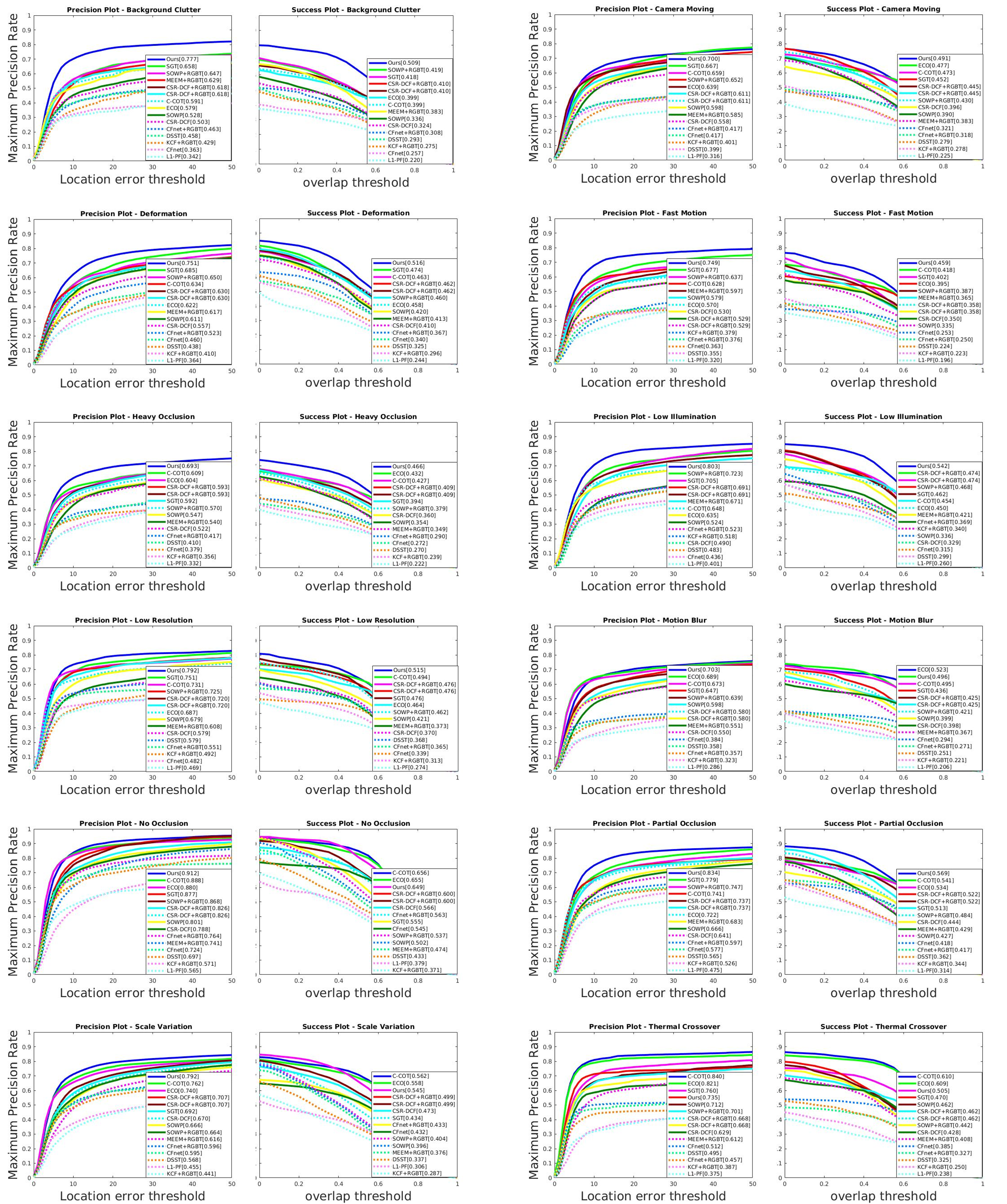}
\caption{The tracking results under various challenging factors. }
\label{resultsAttributes}
\end{figure*}

We also conduct experiments to check the robustness of our proposed RGBT tracker under each challenging factors on the RGBT-234 dataset. The detailed tracking results are shown in Fig. \ref{resultsAttributes}. It is easy to find that our tracker is more robust to most of the challenging factors than the compared algorithms. 

%\begin{figure*}[t]
%\center
%\includegraphics[width=7in]{result_fig_2}
%\caption{The tracking results of our tracker and other state-of-the-art RGBT tracking algorithm. Our tracking results is shown in {\color{green} green}, and the ground truth is the {\color{yellow} yellow} bounding box. }
%\label{visualizationResultsMore}
%\end{figure*} 	

\section{Conclusion}
Many existing RGB-T tracking algorithms directly fuse the image features and train a binary classifier. However, they ignore the attention of the classifier and their local search strategy in tracking-by-detection framework also limit their robustness. In this paper, we propose a reciprocative learning algorithm and use common attention maps as a regularization term to train a more discriminative classifiers. In addition, we also introduce multi-modal target-driven global attention network for high-quality global proposal generation for visual tracking. Extensive experiments on two RGB-T tracking dataset validated the effectiveness of our RGB-T tracker.

\section{Acknowledgements}	
This work is jointly supported by National Natural Science Foundation of China (61702002, 61671018, 61872005), Key International Cooperation Projects of the National Foundation (61860206004), Natural Science Foundation of Anhui Province (1808085QF187), Natural Science Foundation of Anhui Higher Education Institution of China (KJ2017A017), Institute of Physical Science and Information Technology, Anhui University.

\bibliographystyle{IEEEbib}
\bibliography{reference}

\end{document}